\documentclass[10pt,twocolumn,letterpaper]{article}

\usepackage{wacv}
\usepackage{times}
\usepackage{epsfig}
\usepackage{graphicx}
\usepackage{amsmath}
\usepackage{amssymb}
\usepackage{booktabs}
\usepackage{multicol}
\usepackage{multirow}
\usepackage{makecell}
\usepackage{bbm}
\usepackage{algorithm}
\usepackage{algpseudocode}

\ifwacvfinal
\usepackage[breaklinks=true,bookmarks=false]{hyperref}
\else
\usepackage[pagebackref=true,breaklinks=true,colorlinks,bookmarks=false]{hyperref}
\fi

\usepackage[capitalize]{cleveref}
\crefname{section}{Sec.}{Secs.}
\Crefname{section}{Section}{Sections}
\Crefname{table}{Table}{Tables}
\crefname{table}{Tab.}{Tabs.}
\crefname{figure}{Fig.}{Figs.}
\Crefname{figure}{Fig.}{Figs.}

\newcommand{\nj}[1]{\textcolor{black}{#1}}

\floatname{algorithm}{Procedure}

\def\wacvPaperID{1914} 

\wacvalgorithmstrack   

\wacvfinalcopy 

\begin{document}

\title{Leveraging Off-the-shelf Diffusion Model for Multi-attribute \\Fashion Image Manipulation}

\author{Chaerin Kong$^1$\thanks{Work done during internship at NAVER.} \hspace{2mm} DongHyeon Jeon$^2$ \hspace{2mm} Ohjoon Kwon$^2$ \hspace{2mm} Nojun Kwak$^1$\\
$^1$Seoul National University \hspace{2mm} $^2$NAVER\\
{\tt\small veztylord@snu.ac.kr \hspace{1mm} \{donghyeon.jeon, ohjoon.kwon\}@navercorp.com \hspace{1mm} nojunk@snu.ac.kr}
}


\maketitle
\thispagestyle{empty}

\begin{abstract}
   Fashion attribute editing is a task that aims to convert the semantic attributes of a given fashion image while preserving the irrelevant regions. Previous works typically employ conditional GANs where the generator explicitly learns the target attributes and directly execute the conversion. These approaches, however, are neither scalable nor generic as they operate only with few limited attributes and a separate generator is required for each dataset or attribute set. Inspired by the recent advancement of diffusion models, we explore the classifier-guided diffusion that leverages the off-the-shelf diffusion model pretrained on general visual semantics such as Imagenet. In order to achieve a generic editing pipeline, we pose this as multi-attribute image manipulation task, where the attribute ranges from item category, fabric, pattern to collar and neckline. We empirically show that conventional methods fail in our challenging setting, and study efficient adaptation scheme that involves recently introduced attention-pooling technique to obtain a multi-attribute classifier guidance. Based on this, we present a mask-free fashion attribute editing framework that leverages the classifier logits and the cross-attention map for manipulation. We empirically demonstrate that our framework achieves convincing sample quality and attribute alignments.
\end{abstract}

\section{Introduction}

Denoising diffusion models~\cite{ho2020denoising, song2020denoising, dhariwal2021diffusion, ramesh2022hierarchical, saharia2022photorealistic} have recently gained great attention from the research community for their impressive synthesis quality, training stability and scalability. They have demonstrated promising performances across diverse tasks and benchmarks spanning unconditional image synthesis~\cite{dhariwal2021diffusion}, text-driven image generation~\cite{ramesh2022hierarchical, saharia2022photorealistic, nichol2021glide}, image manipulation~\cite{avrahami2022blended, hertz2022prompt} and video synthesis~\cite{ho2022video}. Nevertheless, studies on diffusion models are far from complete; unlike traditional generative model families such as Generative Adversarial Networks (GANs) and Variational Autoencoders (VAEs), the true potential of diffusion models are yet to be fully disclosed. 

One of the reasons for the popularity of diffusion models is their natural capacity to incorporate conditioning information into the generative process. 
Conditional diffusion models typically rely on classifier-guidance~\cite{dhariwal2021diffusion} or classifier-free-guidance~\cite{ho2022classifier}, where the former requires a separately trained classifier (independent of the diffusion model) while the latter involves condition-aware training of the diffusion model from the beginning. Classifier-guidance, in particular, provides a means to leverage the off-the-shelf diffusion model trained on general visual semantics such as Imagenet~\cite{deng2009imagenet}, which can be particularly useful for domains with insufficient public data.~\footnote{Classifiers are generally easier and more straightforward to train or finetune compared to generative models under limited data.}

Fashion domain, being the heart of modern e-commerce, has huge practical upside but retains relatively little publicly available data due to privacy and proprietary issues. To make things worse, the visual semantics are significantly distant from the common generative benchmarks such as Imagenet or FFHQ~\cite{karras2019style}, strongly demanding an adequate adaptation procedure. For these reasons, the field of fashion has been relatively less explored in the deep learning community despite its industrial values.

Image manipulation is a generative task that aims to control the semantics of an input image while preserving the irrelevant details. Fashion image manipulation has great applications in fashion design, interactive online shopping, and personalized marketing. \nj{T}hus several works~\cite{ak2019attribute, ping2019fashion, kwon2022tailor} have posed this task as fashion attribute editing task, \textit{i.e.,} attribute-guided image manipulation, and delivered promising results. However, their real world applications are significantly restricted as (1) they train a separate generative model (typically GAN) for each fashion dataset, and (2) their editing operations are limited to few predefined attributes \nj{such as} color or sleeve length, as the \textit{generative model} has to learn to convert these attributes during training.
As these limitations render them short for a generic and scalable editing system,
we propose to employ an off-the-shelf diffusion model trained on \nj{a} general semantic domain such as Imagenet, and guide its editing with a domain-specific classifier. To the best of our knowledge, this is the first attempt to introduce diffusion models to the fashion domain, especially fashion attribute editing.

This approach has clear advantage over the prior art for several reasons. First, training a classifier is generally much simpler and easier than training a generative model under limited data. As there is clear shortage of well-annotated fashion images, we present an efficient finetuning strategy that empowers the classifier to reason a wide range of fashion attributes at once with the help of recently proposed attention-pooling technique~\cite{yu2022coca}.

Second, as the capacity to understand different fashion attributes has been transferred to the classifier, our manipulation framework can operate in a much greater scale, covering attributes like item category, neckline, fit, fabric, pattern, sleeve length, collar and gender with a single model. This is in clear contrast to previous works that supports only a few editing operations \nj{with separate models}. We later show that training attribute-editing GANs with such a wide set of attributes leads to total training collapse (Fig.~\ref{fig:gan_failure}). Last, we can edit multiple attributes at once in an integrated manner, since we use the guidance signal of a multi-attribute classifier. This is particularly important for fashion domain as various attributes must be in harmony with \nj{one another} to yield an attractive output.

Our fashion attribute classifier adopts a pretrained ViT backbone and a finetuned attention pooling layer in order to best perform multi-attribute classification with relatively small training dataset. 
We use the gradient signal of this classifier to guide the diffusion process as done in \cite{avrahami2022blended, dhariwal2021diffusion}, thus we can alter more than one attributes at once.
For local editing of images, using a user-provided mask that explicitly marks the area to be edited is the most straightforward and widely used approach~\cite{avrahami2022blended}.
However, we leverage the natural capacity of our attribute classifier to attend to the relevant spatial regions during classification, and use the attention signal to suppress \textit{excess modifications} of the original image. This frees the users from the obligation to designate the specific region of interest and simplifies the image manipulation pipeline.

In sum, our contributions can be summarized as

\begin{itemize}
    \item We introduce classifier-guided diffusion as a simple yet generic and effective framework for fashion attribute editing.
    \item We empirically present an efficient finetuning scheme to adapt a pretrained ViT for a domain-specific multi-attribute classification setting.
    \item We demonstrate the effectiveness of our framework with thorough evaluations.
\end{itemize}

\section{Related Works}

\noindent
\nj{\textbf{Diffusion Models}}
~\cite{sohl2015deep, ho2020denoising} are a family of generative models that convert a Gaussian noise into a natural image through an iterative denoising process, which is typically modeled with a learnable neural network such as U\nj{-}Net~\cite{ronneberger2015u}. They have \nj{gained} the attention from both the research community and the public with their state-of-the-art performances in likelihood estimation~\cite{ho2020denoising, nichol2021improved} and sample quality~\cite{dhariwal2021diffusion, saharia2022photorealistic, ramesh2022hierarchical}. Specifically, they have demonstrated impressive results in conditional image synthesis, such as class-conditional~\cite{dhariwal2021diffusion}, image-conditional~\cite{lugmayr2022repaint} and text-conditional~\cite{saharia2022photorealistic, ramesh2022hierarchical} settings. Conditioning a diffusion model is typically done with either the classifier-guidance~\cite{dhariwal2021diffusion} or the classifier-free-guidance~\cite{ho2022classifier}, and \nj{conditional diffusion models have} been shown to be capable of learning an extremely rich latent space~\cite{nichol2021glide, ramesh2022hierarchical}. Recently, a line of works~\cite{song2020denoising, rombach2022high} focus on improving the sampling speed of diffusion model, by either altering the Markovian noising process or embedding the diffusion steps into a learned latent space. Another group~\cite{kim2021diffusionclip, avrahami2022blended, hertz2022prompt} studies the applications of diffusion models such as text-guided image manipulation. 

\noindent
\textbf{Image Manipulation}
\nj{or i}mage editing has been a long standing challenge in the computer vision community with a wide range of practical applications~\cite{zhu2017unpaired, patashnik2021styleclip, kim2019u, kong2021smoothing}. Image manipulation with deep models typically accompany editing operations in the latent space to produce semantic and natural modifications. Hence, image manipulation with Generative Adversarial Networks (GANs) pose\nj{s} a new problem of GAN inversion~\cite{tov2021designing, richardson2021encoding}, which is the initial step to find a latent vector corresponding to the image that needs to be altered. Recently, as diffusion models rise as prominent alternatives, image manipulation using diffusion models are widely being studied. Blended-diffusion~\cite{avrahami2022blended} uses a user-provided mask and a textual prompt during the diffusion process to \textit{blend} the target and the existing background iteratively. A concurrent work of ours, prompt-to-prompt~\cite{hertz2022prompt} captures the text cross-attention structure to enable purely prompt-based scene editing \nj{without any explicit masks}. In our work, we take blended-diffusion as the starting point, and incorporate a domain-specific classifier and its attention structure for mask-free multi-attribute fashion image manipulation.

\noindent
\textbf{Fashion Attribute Editing}
\nj{is a highly practical task for which several works have been studied.} 
AMGAN~\cite{ak2019attribute} uses Class Activation Map and two discriminators that identifies both real/fake and the attributes. Fashion-AttGAN~\cite{ping2019fashion} improves upon AttGAN~\cite{he2019attgan} to better fit the fashion domain, by including additional optimization objectives. VPTNet~\cite{kwon2022tailor} aims to handle larger shape changes by posing attribute editing as \nj{a} two-stage procedure of shape-then-appearance editing. These works all employ \nj{the} GAN framework, and thus the range of attributes supported for editing is relatively limited. We explore fashion attribute editing with an off-the-shelf diffusion model that has learned a rich semantic latent space from a set of general images to simplify the generative pipeline and achieve more generic editing capabilities.

\section{Approach}

\subsection{Preliminaries}

\noindent
\textbf{Denoising Diffusion Probabilistic Models} 
(DDPMs)~\cite{ho2020denoising} learn to invert the Markovian noising process \nj{which is} typically parameterized by isotropic Gaussian. They were shown to achieve the best performance under the reweighted objective~\cite{ho2020denoising}, and have demonstrated state-of-the-art performances on a wide range of generative benchmarks~\cite{dhariwal2021diffusion, saharia2022photorealistic}. We present a brief overview of diffusion models below.

Given a clean data distribution $x_0 \sim q(x_0)$, a Markovian noising process can be defined as:
\begin{align}
    &q(x_t|x_{t-1}) = \mathcal{N}(\sqrt{1-\beta_t}x_{t-1}, \beta_t \mathbf{I})\\
    &q(x_1,...,x_T|x_0) = \prod_{t=1}^{T}q(x_t|x_{t-1}),
\end{align}
with $\beta_t$ being the noise scale controlling the diffusion process \nj{and $\mathbf{I}$ being the identity  matrix}.
It is shown that with a large enough total time steps $T$, $x_T$ can be regarded as an isotropic Gaussian variable. As we are gradually adding Gaussian noises, an arbitrary noising step can be directly computed as
\begin{equation}
    q(x_t|x_0) = \mathcal{N}(\sqrt{\Bar{\alpha}_t}x_0, (1-\Bar{\alpha}_t)\mathbf{I}),
\end{equation}
\nj{with a scalar $\Bar{\alpha}_t = \prod_{i=1}^t (1-\beta_i)$.}

To sample under the diffusion framework, we \textit{reverse} the forward denoising process. In other words, we begin with a Gaussian noise $x_T \sim \mathcal{N}(0, \mathbf{I})$, and go through a series of posterior sampling steps, $x_{t-1} \sim q(x_{t-1}|x_t)$, which is typically modeled with a neural network with learnable parameters $\theta$\nj{:}
\begin{equation}
    p_\theta(x_{t-1}|x_t) = \mathcal{N}(\mu_\theta(x_t, t), \Sigma_\theta(x_t, t)).
\end{equation}

\cite{ho2020denoising} shows that with mild assumptions, this posterior can be represented as a \nj{Gaussian distribution with a} diagonal \nj{covariance} and the mean $\mu_\theta(x_t, t)$ can be computed from $\epsilon_\theta(x_t, t)$. Thus, the neural network is trained to predict the noise $\epsilon_\theta(x_t,t)$ instead of the mean under the following reconstruction objective:
\begin{equation}
    \mathcal{L} = ||\epsilon_\theta(x_t, t)-\epsilon||^2.
\end{equation}

Several works~\cite{nichol2021improved, dhariwal2021diffusion} recently introduce empirical strategies to produce samples with better quality, such as model architecture and objective weighting. 

\hfill \break
\noindent
\textbf{Blended-Diffusion}~\cite{avrahami2022blended} adapts the diffusion model for \nj{the} text-driven image manipulation task. Employing a user-provided regional mask $M$, a pretrained vision-language encoder, \textit{e.g.,} CLIP~\cite{radford2021learning}, provides guidance to form a text-aligned visual semantic in the masked region. At each diffusion step, the masked region guided by CLIP and the unmasked background region are \textit{blended} to form a smooth and natural visual composition. To evade adversarial effects stemming from ascending gradients of CLIP logits, the authors propose \textit{Extending Augmentations} that create multiple views of a given image. 

In this work, we propose to tailor this generic editing framework for fashion domain with concise modifications. We first show that naively applying this general framework yields suboptimal results, then present \nj{an} efficient finetuning technique to prepare a multi-attribute fashion classifier that is capable of reasoning different spatial regions for classification of different attributes. Lastly, we leverage its natural attention map for mask-free fashion attribute editing, completing an extremely simple and generic fashion image manipulation framework. \nj{The overall procedure of ours is described in Procedure~\ref{alg:cap}.}



\begin{figure}[t]
\centering
  \includegraphics[width=0.45\textwidth]{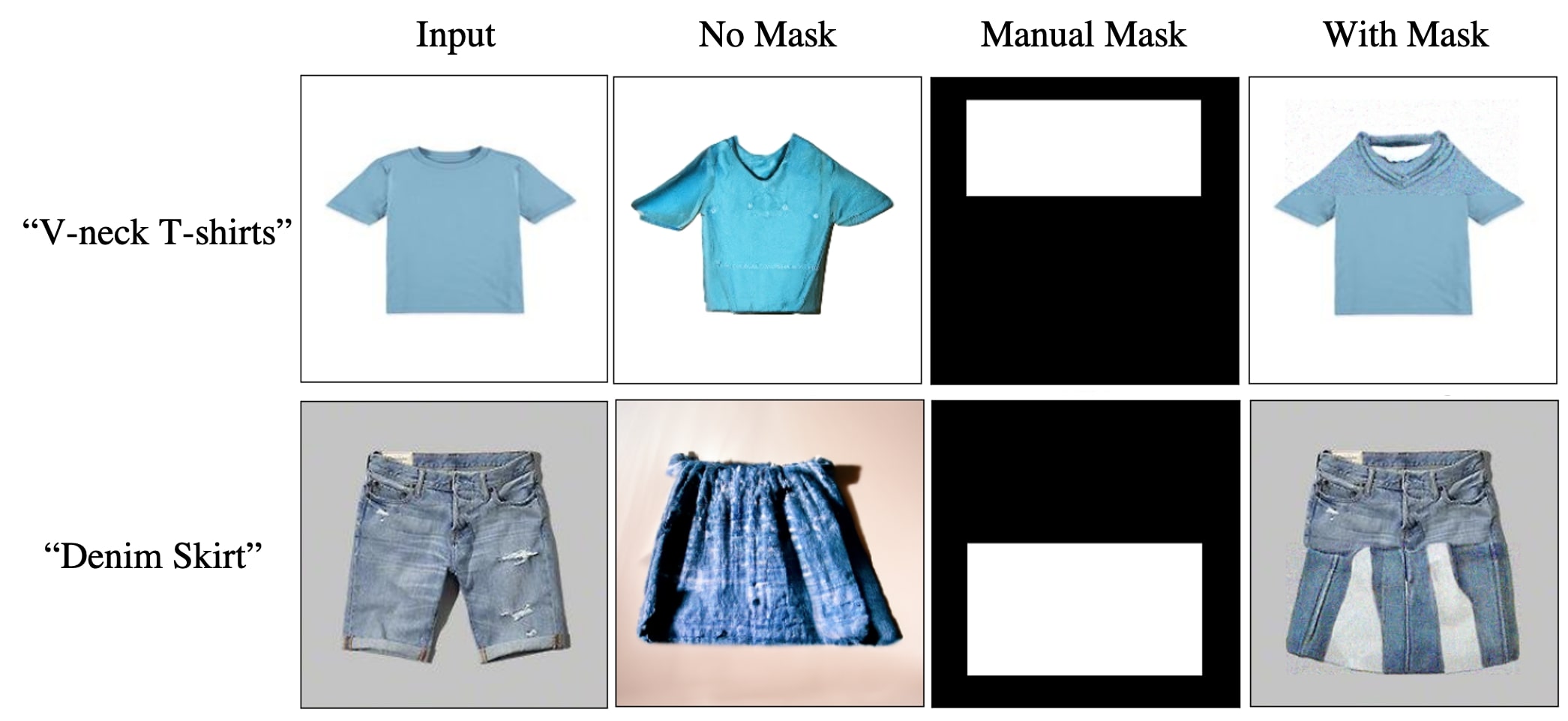}
  \caption{Naive application of blended-diffusion in fashion domain leads to suboptimal results.}
  \label{fig:bd_failure}
\end{figure}

\begin{algorithm}
\caption{Our proposed fashion attribute editing pipeline, with the off-the-shelf diffusion model $\theta$ and the attribute classifier $f$.}\label{alg:cap}
\begin{algorithmic}
\Require original image $X$, target attributes $A$, diffusion step $k$, number of augmentation views $N$ and attribute loss weight $\lambda$
\Ensure edited output $\hat{X}$
\State $X_k \sim \mathcal{N}(\sqrt{\Bar{\alpha}_k}X_0, (1-\Bar{\alpha}_k)\mathbf{I})$
\For{$t$ in $k$:$0$}
    \State $\hat{X}_0 \gets {{X_t - \sqrt{1-\Bar{\alpha}_t}\epsilon_\theta(X_t, t)}\over{\sqrt{\Bar{\alpha}_t}}}$
    \State $\hat{X}_{0,aug} \gets ExtendingAugmentations(\hat{X}_0, N)$
    \State $\nabla_{attr} \gets {1 \over N}\sum_{i=1}^N\nabla_{\hat{X}_{0,aug}}f(\hat{X}_{0,aug}, A)$
    \State $\Bar{M} \gets {1 \over |A|} \sum_{j=1}^{|A|} AttentionMap(f, \hat{X}_{0}, A)$
    \State $\nabla_{bckg} \gets -\nabla_{\hat{X}_{0}}\mathcal{L}_{bckg}(X, \hat{X}_{0}, \Bar{M})$
    \State $\nabla_{total} \gets \lambda \nabla_{attr} + \nabla_{bckg}$
    \State $X_{t-1} \sim \mathcal{N}(\mu_\theta(X_t)+ \Sigma_\theta(X_t)\nabla_{total},  \Sigma_\theta(X_t)) $
\EndFor
\State \textbf{return} $X_{-1}$

\end{algorithmic}
\end{algorithm}

\begin{figure*}[t]
\centering
  \includegraphics[width=0.95\textwidth]{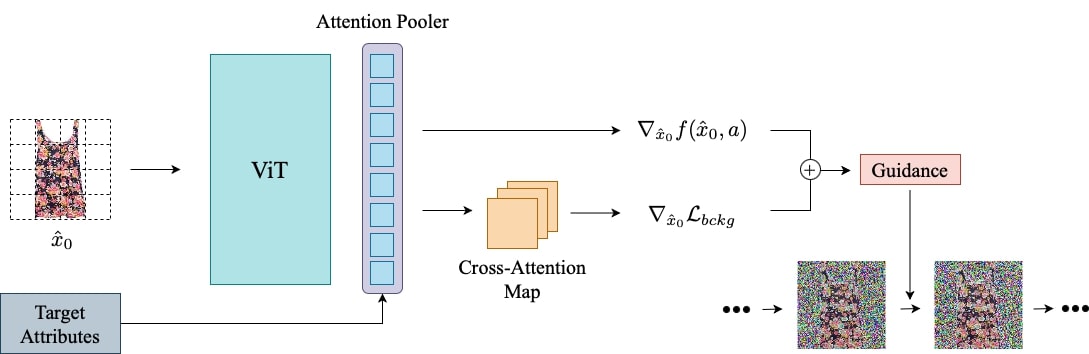}
  \caption{Overall pipeline of our proposed framework. A predicted clean image is first patch-embedded as input to the ViT backbone. The attention-pooling layer on top of the ViT returns attribute logits and the cross-attention maps, which are used to guide the diffusion steps.}
  \label{fig:qual}
\end{figure*}

\subsection{Taming ViT for Multi-attribute Guidance}

\nj{As shown in Fig.~\ref{fig:bd_failure}, o}ur early experiments show that naively applying the guidance signal from a pretrained model (\textit{e.g.,} CLIP) as in blended-diffusion yields unsatisfactory results in fashion domain due to the severe domain gap. To provide a more fine-grained guidance to our diffusion model, we explore ways to efficiently train a domain-specific classifier.

Vision Transformers (ViTs)~\cite{dosovitskiy2020image} have recently shown impressive performances across diverse computer vision tasks, frequently replacing the state-of-the-arts previously held by convolutional neural networks~\cite{chen2021empirical, liu2021swin}. In conventional ViTs, the global semantic of an input image is aggregated to a single \texttt{[CLS]} token, which is commonly projected with additional learnable layers for downstream tasks. However, in \nj{our} multi-attribute classification setting, 
this can lead to \textit{over-simplification} of visual semantics, as different attributes reside in different spatial regions. In other words, we have to look at different parts of a clothing to determine \nj{different attributes such as} neckline or sleeve length. Hence, we adopt recently proposed attention-pooling mechanism \cite{yu2022coca} to aggregate the spatial features in a learnable way with an additional cross-attention layer. This enables our model to attend to different spatial regions for classification of different attributes. After the attention pooling, each attribute token is projected for classification, and they are optimized with the classic cross-entropy loss \nj{formulated as follows:}
\begin{equation}
    \mathcal{L}_{cls} = -\sum_{i=1}^{|A|} \log{exp(x_{target}^{(i)}) \over {\sum_{j=1}^{C^{(i)}} exp(x_{j}^{(i)})}} \cdot \mathbbm{1}(y^{(i)} \neq -1).
\end{equation}
\nj{Here, $|A|$ is the number of attributes}, $x^{(i)}$ refers to the model prediction for \nj{the} $i$-th attribute, $C^{(i)}$ is the number of classes for the attribute, and $y^{(i)}$ indicates the ground truth label. We mask those with missing attributes, \textit{e.g.,} neckline for a skirt. Note that we drop the mini-batch notation for brevity.

As ViTs are famous for being notoriously data-hungry~\cite{dosovitskiy2020image}, we choose to finetune a ViT pretrained on a larger dataset. We explore different initialization and finetuning methods and present the best practice for our setting in Sec.4.1. 


\subsection{Local Image Editing with Patch-level Attention}

One of the challenges of image editing is to concisely alter the regions to suit the given condition while keeping the irrelevant parts unchanged. Instead of relying on a user-provided mask \nj{as done in \cite{avrahami2022blended}}, we employ the attention map of our classifier to determine the relevant regions. This is distinguished from other works that use separate procedures such as Class Activation Map~\cite{zhou2016learning} or Grad-CAM~\cite{selvaraju2017grad}, as we simply use the attention map of each attribute token (in the attention pooling layer), which is computed during the classifier forward pass. 

Formally, for a multi-attribute image editing, we average-pool the spatial attention maps for the target attribute tokens, and impose background preservation loss on the low-attention regions as \nj{follows:}
\begin{equation}
    \mathcal{L}_{bckg} = (1-\Bar{M}) \odot ||\hat{X}-X||_2 + (1-\Bar{M}) \odot \nj{pd}(\hat{X}, X),
\end{equation}
where $\Bar{M}$ refers to the average-pooled spatial attention maps, $\odot$ indicates the Hadamard product, $\hat{X}$ is the predicted output from $X$, and \nj{$pd$} is Learned Perceptual Image Patch Similarity \nj{(LPIPS)}~\cite{lpips}, also known as the perceptual distance.

\subsection{Overall Framework}

We build our framework upon \textit{blended-diffusion}~\cite{avrahami2022blended}, and make modifications to suit the off-the-shelf diffusion model for fashion attribute editing. First, we replace CLIP guidance with our finetuned classifier guidance that supports fine-grained multi-attribute editing. Then, we \nj{get rid of} the user mask and enforce background preservation with the classifier attention map. We go through the diffusion steps with the initial input image, where the classifier guidance signal pulls the diffusion process towards the target attributes. The overall pipeline is illustrated in \cref{fig:qual}.

\section{Experiments}

\subsection{Datasets and Baselines}

We use the commonly used Shopping100k~\cite{ak2018efficient} dataset for training and evaluations. As our diffusion model does not need additional training, we only train the classifier on this dataset. For baselines, we compare our method with StarGAN~\cite{choi2018stargan} and Fashion-AttGAN~\cite{ping2019fashion}, two representative attribute manipulation methods of which the latter is further suited for the fashion domain. We train these models on Shopping100k with 8 most common attributes, summing to 100 labels total. The details are specified in \cref{tab:attrs}.

\begin{table}[h]
\centering
\resizebox{0.98\columnwidth}{!}{
\begin{tabular}{c|c|l}
\Xhline{4\arrayrulewidth}
Attribute     & \#Classes & \multicolumn{1}{c}{Class}                                                                                                                                                                               \\ \hline
Category      & 16        & \begin{tabular}[c]{@{}l@{}}Coat, Jacket, Suit, Shirt, T-shirt, Jumper, Shorts\\ Trouser, Jean, Swimming, Jumpsuit, Pyjamas, \\ Tracksuit, Bottoms, Tracksuit, Skirt, Dress\end{tabular}                 \\
Collar        & 17        & \begin{tabular}[c]{@{}l@{}}Buttondown, Cutaway, High, Hood, Kent, Lapel,\\ Lined, Mandarin, Polo, Round, Shawl, Turndown,\\ V-neck, Peter Pan, Volant, Shirt, Chin\end{tabular}                         \\
Fabric        & 14        & \begin{tabular}[c]{@{}l@{}}Canvas, Crocheted, Denim, Fleece, Hardshell,\\ Jersey, Jersey Lace, Lace, Mesh, Mesh Jersey,\\ Rib, Softshell, Sweat, Leather\end{tabular}                                   \\
Fit           & 15        & \begin{tabular}[c]{@{}l@{}}Bootcut, Flared, High waist, Jeggings, Large, \\ Loose, Low, Oversize, Regular, Skinny, Slim, \\ Small, Straight, Tailered, Tapered\end{tabular}                             \\
Gender        & 2         & Male, Female                                                                                                                                                                                            \\
Neckline      & 11        & \begin{tabular}[c]{@{}l@{}}Boat, Backless, Cache-coeur, Henley, Low v-neck, \\ Low round, Off-the-shoulder, Round, Square, \\ V-neck, Envelope\end{tabular}                                             \\
Pattern       & 16        & \begin{tabular}[c]{@{}l@{}}Animal, Burnout, Camouflage, Checked, Marl,\\ Color gradient, Colorful, Floral, Herringbone, \\ Paisley, Photo, Pinstriped, Plain, Polka dot, \\ Print, Striped\end{tabular} \\
Sleeve Length & 9         & \begin{tabular}[c]{@{}l@{}}3/4, Spaghetti, Sleeveless, Elbow, Extra long, \\ Extra short, Long, Short, Strapless\end{tabular}                                                                           \\ \hline
\end{tabular}}
\vspace{2mm}
\caption{Details on the attributes and their corresponding classes. We use the attribute annotations of Shopping100k dataset.}
\label{tab:attrs}
\end{table}

\subsection{Implementation Details}

We leverage the unconditional diffusion model pretrained on 256x256 scale Imagenet~\cite{dhariwal2021diffusion}. Surprisingly enough, despite the fact that the diffusion model has never seen any fashion-specific images, \nj{the} classifier alone can guide fashion attribute editing in the rich latent space of the diffusion model. For the classifier, we test three initialization methods: random, Imagenet ViT and CLIP ViT. ViT-Large model was used for all experiments, and we added an attention pooling layer on top of the final transformer block. This attention pooler is essentially a cross-attention layer with 8 queries, each corresponding to an attribute. These tokens are linearly projected for final classification.

\subsection{Collapse of Attribute-editing GANs under Wider Attribute Space}

Previous methods using attribute-editing GANs typically focus on one or two attributes (\textit{e.g.,} color or sleeve length), as these methods enforce the generator itself to learn to reason about different fashion attributes. 
Hence, we first observe how these conventional GANs perform in the more generic manipulation setting, where we tackle most of the widely used fashion attributes, ranging from item category to fit and neckline, with a single framework.

\begin{figure}[t]
\centering
  \includegraphics[width=0.48\textwidth]{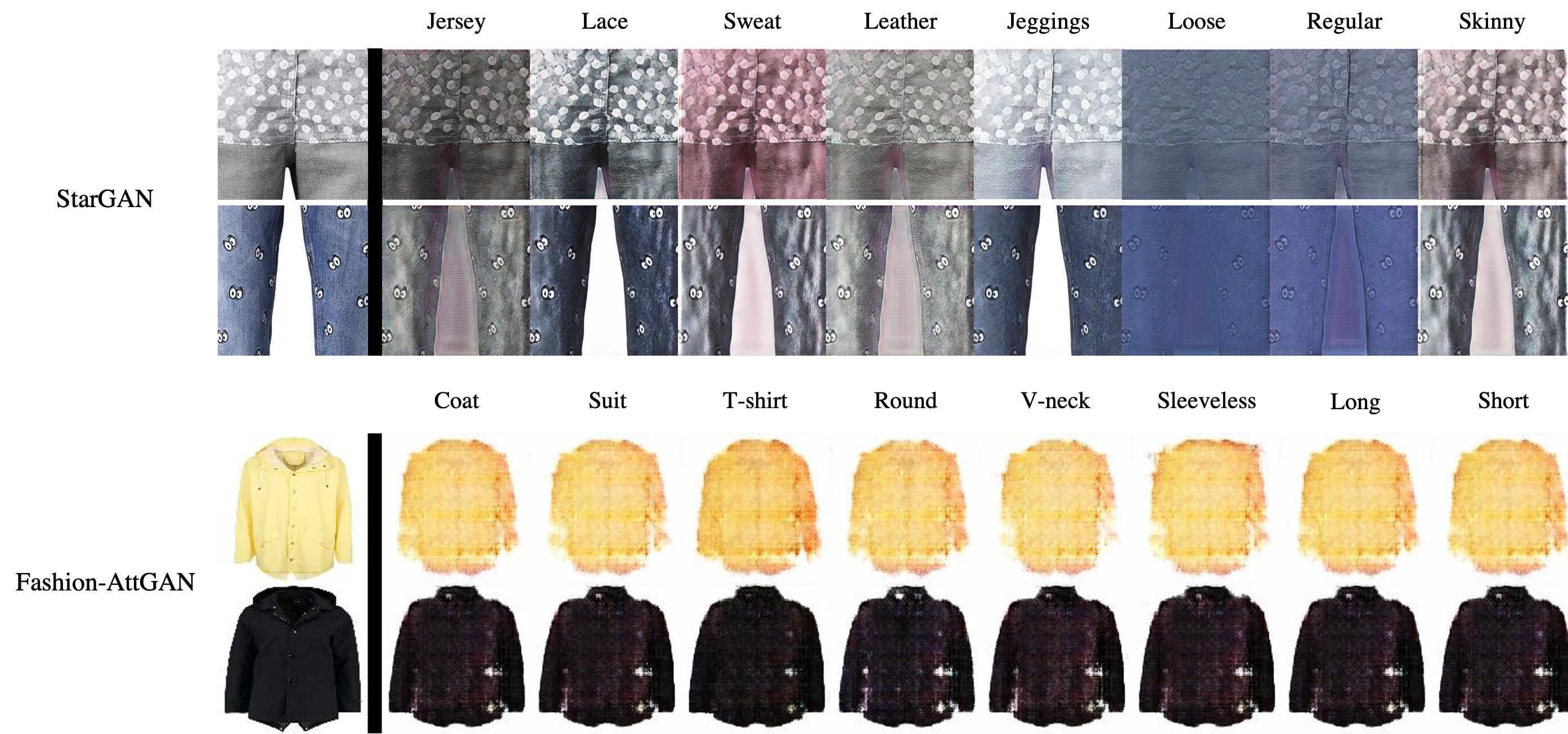}
  \vspace{0.5mm}
  \caption{Attribute-editing GANs fail to handle generic manipulation scenario with more attributes and display training collapse.}
  \label{fig:gan_failure}
\end{figure}

\cref{fig:gan_failure} clearly illustrates that naively applying attribute-editing GANs~\cite{choi2018stargan, ping2019fashion} leads to severe training collapse. As these methods aim to embed attribute manipulation capacity into the \textit{generator}, they have inherent disadvantage for scalability. For multi-attribute editing, \textit{e.g.,} editing the neckline \textit{and} the sleeve length, multiple generative models have to be trained to perform a sequence of manipulation operations, which greatly limits their practical applications. 

\subsection{ViT Finetuning}

As demonstrated in the previous section, training an attribute-aware generative model is not a scalable option. Therefore, we investigate ways to leverage a well trained generative model with a domain-specific attribute-aware classifier, deferring the attribute reasoning capacity to the classifier for the sake of \nj{a} generic and scalable attribute editing system.

We first present empirical findings from the classifier training. As annotated data is relatively scarce in the fashion domain, we regard the initialization scheme to be a key factor in the final classifier performance. Hence, we explore three settings: (1) random initialization, (2) Imagenet-pretrained initialization and (3) CLIP-pretrained initialization. For the last two, we further compare different finetuning approaches, \textit{i.e.,} how many layers or blocks to set as learnable (unfreeze) during the adaptation period. 

\begin{table*}[t]
\centering
\resizebox{0.95\textwidth}{!}{
\begin{tabular}{llccccccccc}
\Xhline{4\arrayrulewidth}
                    &                     & Category      & Fabric        & Sleeve Length & Pattern       & Gender        & Fit           & Collar        & Neckline      & Average       \\ \hline
Random Init.        &                     &               &               &               &               &               &               &               &               &               \\
                    & End-to-End          &  30.1         &  56.6         &  51.8         &  50.3         &    66.1       & 57.1          &  31.0         &   65.1        & 51.0          \\ \hline
Imagenet-pretrained &                     &               &               &               &               &               &               &               &               &               \\
                    & Attention-Pool Only & 85.4          & 58.7          & 84.8          & 76.1          & 95.0          & 66.4          & \textbf{91.4} & \textbf{84.4} & 80.3          \\
                    & Last2               & 67.0          & 52.8          & 78.9          & 48.5          & 74.6          & 59.7          & 81.6          & 78.1          & 67.6          \\
                    & Last4               & 44.8          & 52.8          & 60.8          & 45.0          & 82.0          & 58.6          & 76.1          & 76.2          & 62.0          \\
                    & Last6               & 43.1          & 52.1          & 73.0          & 44.0          & 77.4          & 54.9          & 76.0          & 71.1          & 61.5          \\ \hline
CLIP-pretrained     &                     &               &               &               &               &               &               &               &               &               \\
                    & Attention-Pool Only & 86.3          & 60.2          & \textbf{84.9} & 80.4          & 95.8          & 69.9          & 91.0          & 83.3          & \textbf{81.5} \\
                    & Last6               & \textbf{87.2} & \textbf{67.9} & 84.4          & \textbf{87.2} & \textbf{97.4} & \textbf{70.9} & 75.3          & 78.0          & 81.0          \\
                    & Last12              & 85.6          & 67.8          & 83.2          & 86.4          & 97.1          & 69.9          & 78.5          & 76.5          & 80.6          \\
                    & Last18              & 82.3          & 66.8          & 79.7          & 83.2          & 95.9          & 68.4          & 72.9          & 73.7          & 77.8          \\
                    & Last24              & 51.7          & 61.9          & 70.4          & 61.5          & 84.7          & 61.9          & 41.6          & 65.1          & 62.3          \\ \hline
\end{tabular}}
\vspace{2mm}
\caption{Quantitative evaluations on different finetuning strategies. We present finetuning results on Shopping100k~\cite{ak2018efficient}, and all models adopt ViT-L for comparison.}
\label{tab:quant1}
\end{table*}

\begin{table*}[t]
\centering
\resizebox{0.95\textwidth}{!}{
\begin{tabular}{llccccccccc}
\Xhline{4\arrayrulewidth}
                    &             & Category      & Fabric        & Sleeve Length & Pattern       & Gender        & Fit           & Collar        & Neckline      & Average       \\ \hline
Imagenet-pretrained &             &               &               &               &               &               &               &               &               &               \\
                    & No Aug.     & \textbf{85.4} & \textbf{58.7} & \textbf{84.8} & \textbf{76.1} & \textbf{95.0} & \textbf{66.4} & \textbf{91.4} & \textbf{84.4} & \textbf{80.3} \\
                    & Random Aug. & 81.5          & 57.3          & 83.3          & 73.7          & 92.7          & 65.2          & 91.2          & 82.0          & 78.4          \\ \hline
CLIP-pretrained     &             &               &               &               &               &               &               &               &               &               \\
                    & No Aug.     & 86.3          & 59.8          & \textbf{85.5} & 79.6          & \textbf{96.6} & 67.7          & 89.5          & 82.5          & 80.9          \\
                    & Random Aug. & 86.3          & \textbf{60.2} & 84.9          & \textbf{80.4} & 95.8          & \textbf{69.9} & \textbf{91.0} & \textbf{83.3} & \textbf{81.5} \\ \hline
\end{tabular}}
\vspace{2mm}
\caption{Ablation on random data augmentations. Only the attention pooling layer is finetuned with Shopping100k dataset. Note that in \cref{tab:quant1}, we report the numbers from \textit{No Aug.} model for Imagenet-pretrained initialization.}
\label{tab:quant2}
\end{table*}

\begin{table*}[t!]
\centering
\resizebox{0.85\textwidth}{!}{
\begin{tabular}{cccccccccc}
\Xhline{4\arrayrulewidth}
Model & Category      & Fabric        & Sleeve Length & Pattern       & Gender        & Fit           & Collar        & Neckline      & Average       \\ \hline
B/32  & 83.9          & 58.4          & 84.3          & 77.9          & 95.0          & 66.0          & 88.4          & 79.5          & 79.2          \\
B/16  & 84.8          & 58.8          & \textbf{85.5} & 78.6          & 95.3          & \textbf{67.8} & 88.6          & 80.2          & 80.0          \\
L/14  & \textbf{86.3} & \textbf{59.8} & \textbf{85.5} & \textbf{79.6} & \textbf{96.6} & 67.7          & \textbf{89.5} & \textbf{82.5} & \textbf{80.9} \\ \hline
\end{tabular}}
\vspace{2mm}
\caption{Model size ablation. We observe steady improvements in the overall performance as the model gets bigger, which agrees with recent findings.}
\label{tab:quant3}
\end{table*}

\cref{tab:quant1} shows the result. We observe that with relatively limited data, freezing most of the parameters while \nj{training} a learnable attention pooling layer yields the best overall performance. This indicates that the pretrained ViT backbone retains a reasonable level of general visual reasoning, and the additional attention pooling layer successfully extracts diverse visual information to perform the fine-grained multi-attribute classification. Moreover, initializing with pretrained weights helps, and CLIP pretraining covers a wider range of visual semantic compared to Imagenet, possibly leading to better outcomes. Lastly, as different attributes demand different aspects of visual reasoning, the classification accuracy trend is not always consistent. We basically choose the model with the highest average score for the best overall guidance.

Data augmentations have been shown to affect the performance of visual classifier models~\cite{steiner2021train}. \nj{Thus,} we explore the influence of data augmentations in our framework. From \cref{tab:quant2}, we gain conflicting insights; random augmentation boosts the performance for CLIP-pretrained ViT but damages it for the Imagenet-pretrained variant. We note that this result is obtained from finetuning the attention pooling layer on Shopping100k dataset, so it is possible to observe different trends in different settings. In our situation, we hypothesize that as we freeze most of the model parameters, imposing strong augmentations could result in \textit{over-pressuring} the limited learnable parameters. As CLIP-pretraining prepares the model for a much greater diversity (compared to Imagenet, in terms of data scale), this potential adversarial effect only realizes for the less robust model variant. 

Lastly, we explore the performance trend depending on the model size. \cref{tab:quant3} shows that bigger model\nj{s} (or longer input sequence) consistently \nj{yield} better outcome\nj{s}, even under \nj{a} strict finetuning scheme. Hence, we adopt the best performing model variant, CLIP-pretrained ViT-L/14, as our diffusion guidance in the following sections.

\begin{figure*}[t]
\centering
  \includegraphics[width=0.95\textwidth]{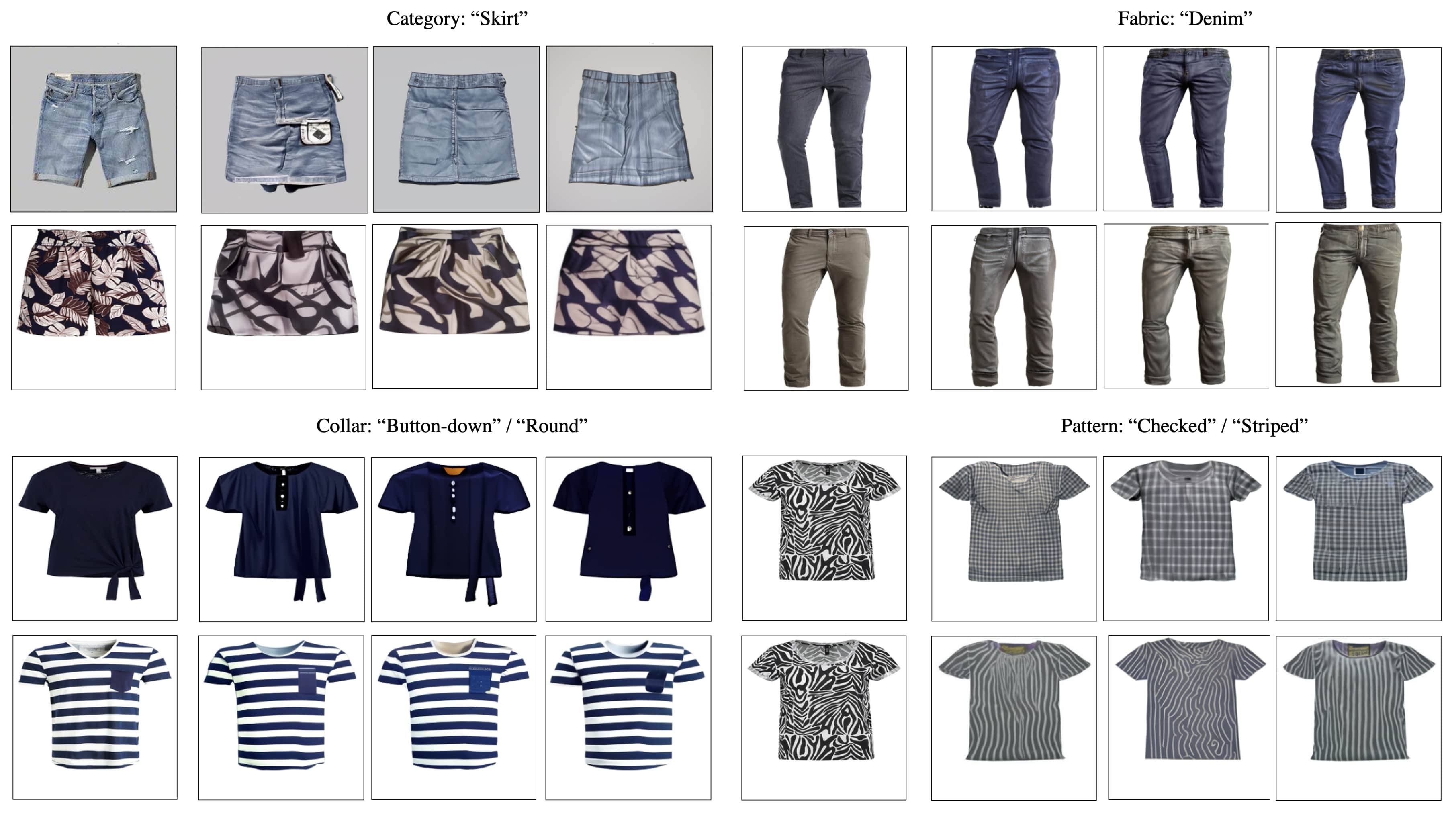}
  \caption{Qualitative evaluations on generic fashion attribute editing. We display the original source images on first and fifth column, and the rest are edited samples. Our framework handles both texture manipulation and shape deformation proficiently, producing plausible samples across various attributes.}
  \label{fig:qual1}
\end{figure*}

\begin{figure}[t]
\centering
  \includegraphics[width=0.95\columnwidth]{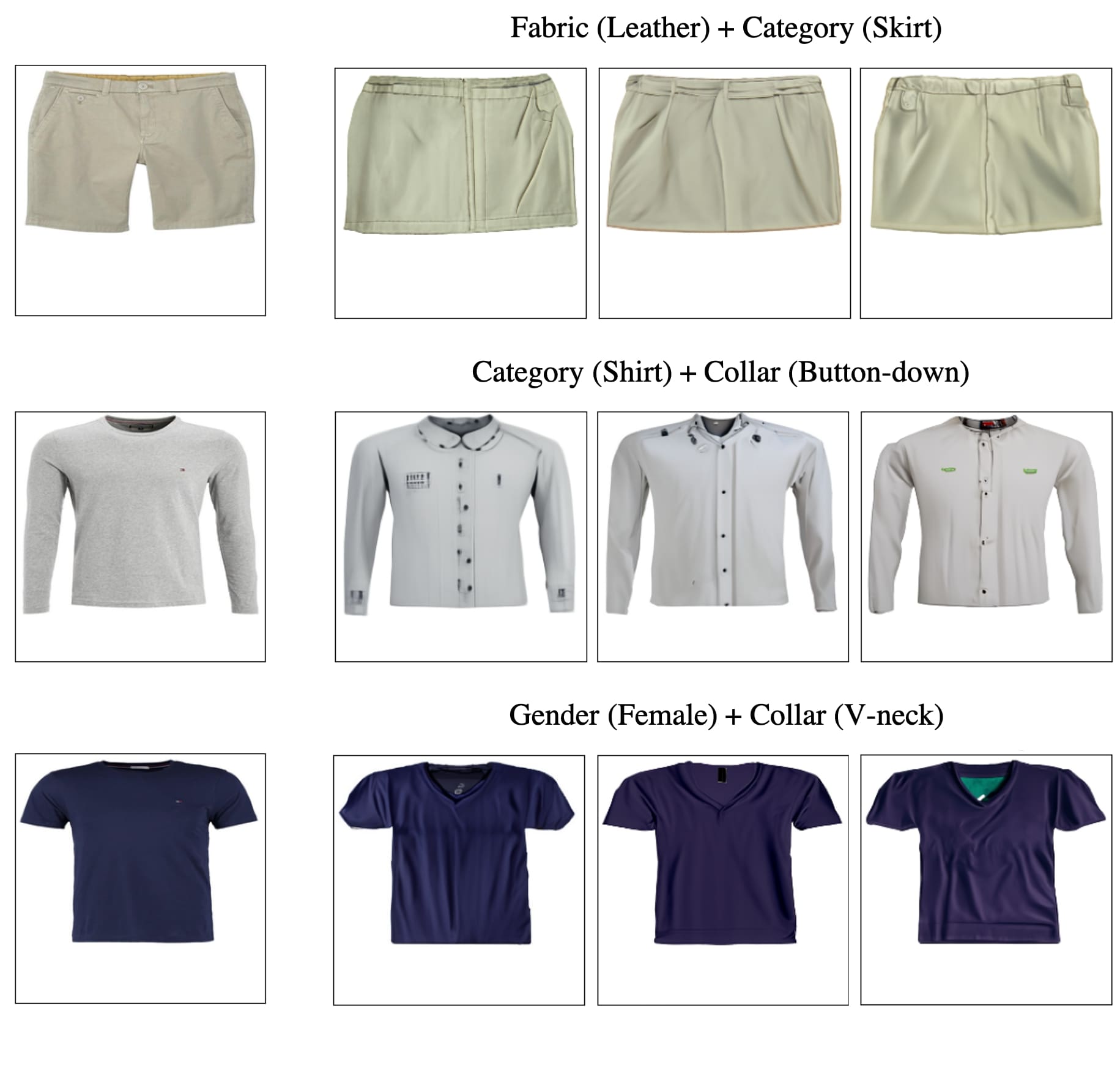}
  \caption{Qualitative results from multi-attribute editing.}
  \label{fig:qual2}
\end{figure}

\begin{figure}[t]
\centering
  \includegraphics[width=0.75\columnwidth]{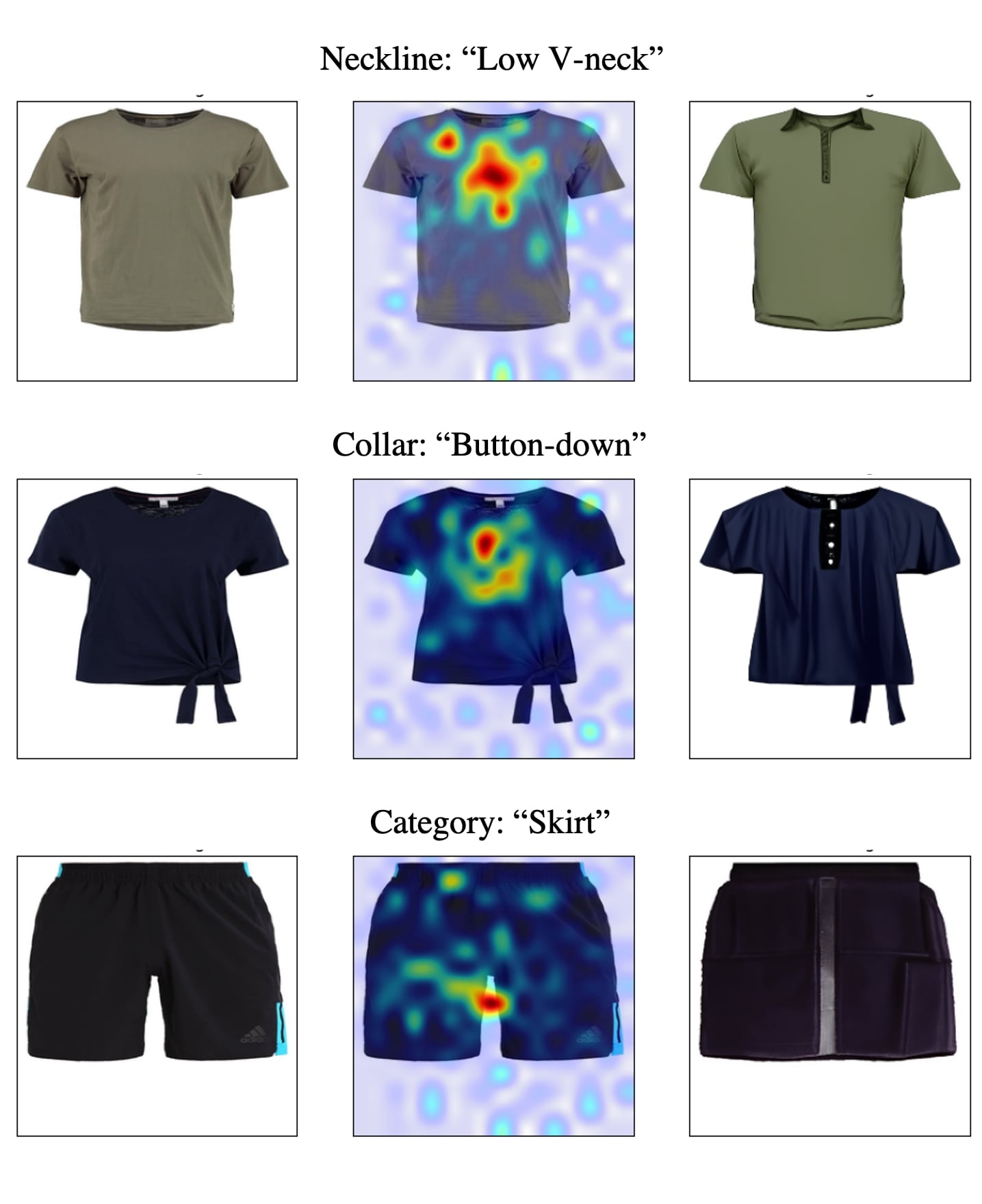}
  \caption{Classifier attention-map visualizations.}
  \label{fig:attn}
\end{figure}

\begin{figure}[t]
\centering
  \includegraphics[width=0.9\columnwidth]{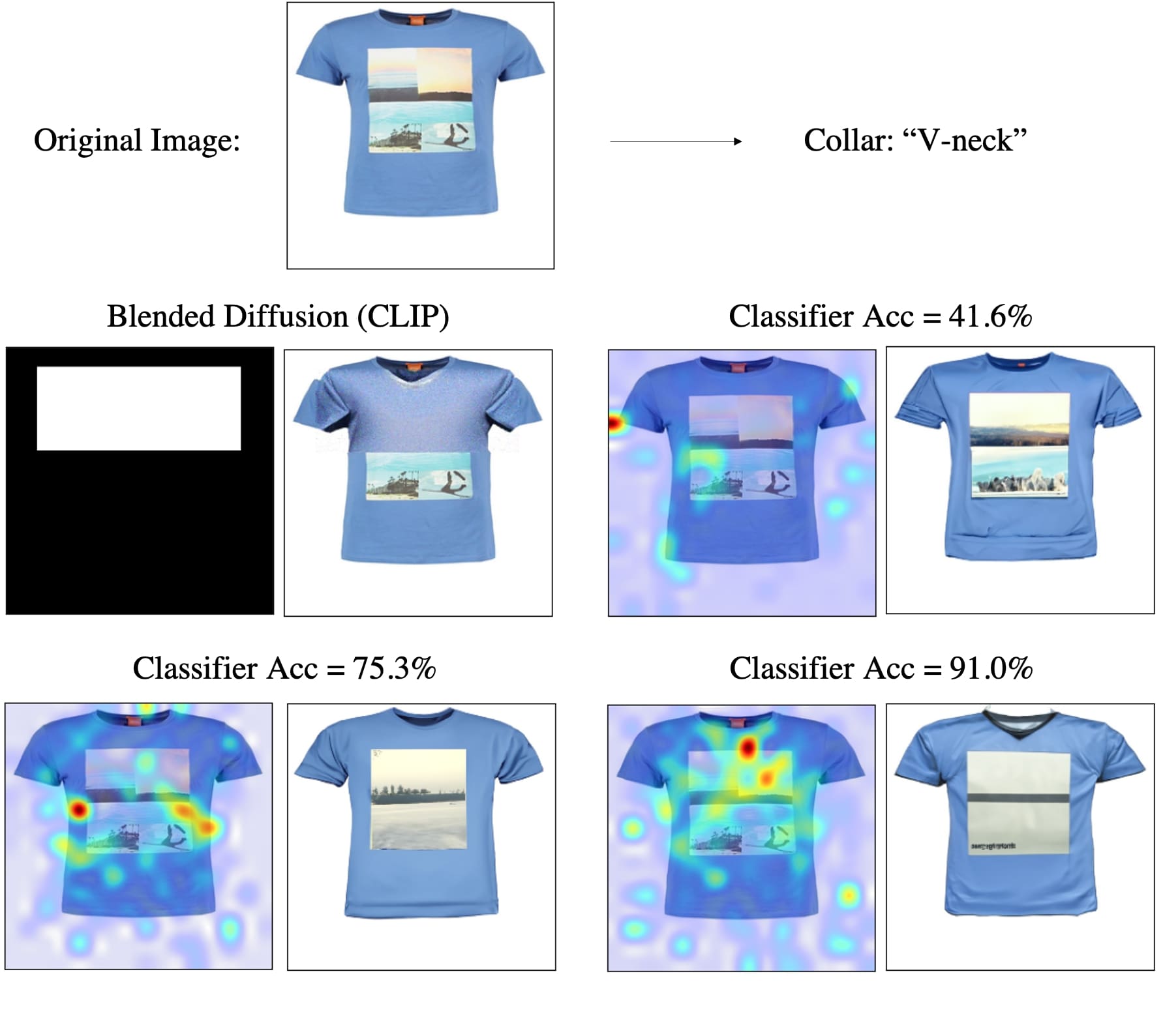}
  \caption{Classification accuracy and the guidance quality. \nj{The b}etter classifier better guides the diffusion model.}
  \label{fig:acc_abl}
\end{figure}

\subsection{Generic Fashion Attribute Editing}

Thanks to the domain-specific multi-attribute classifier, we can leverage a pretrained off-the-shelf diffusion model (not necessarily adapted for fashion domain) for generic fashion attribute editing. \cref{fig:qual1} shows qualitative manipulation results on diverse attributes. We note that these editing operations are all done with a single finetuned classifier and an off-the-shelf diffusion model. Unlike attribute-editing GANs that collapse as the number of attributes increases (see \cref{fig:gan_failure}), our framework shows satisfactory editing \nj{capabilities} 
across various attributes as the classifier can handle a much wider set of attributes with little \nj{in}stability when compared to the generator. Especially, to our knowledge, attributes like item category or fabric have not been directly explored in the previous works on attribute editing mainly due to their difficulties. Since we employ a powerful diffusion model trained with a large dataset, providing the \textit{right} guidance yields impressive outcomes. 

Moreover, it is noteworthy how the diffusion model controls both the texture and the shape in a stable manner. The left side of \cref{fig:qual1} displays shape deformation, and the right side shows texture manipulations, where samples with convincing qualities are consistently delivered. The whole editing pipeline does not require any texture- or shape-specific modules, which is distinguished from the prior art~\cite{kwon2022tailor}, making our framework extremely simple and generic.

In \cref{fig:qual2}, we present multi-attribute editing results, \textit{i.e.,} manipulating multiple attributes at once. This task is apparently more challenging as it requires the editing of different attributes to be in harmony with one another. We observe that our framework produces reasonable outcomes across different attribute combinations, sometimes deforming the original image to a significant extent to meet the complex target condition. 

As our method uses the attribute-wise classifier attention-map for background preservation, the ability to generate adequate spatial attention-maps is critical. \cref{fig:attn} displays the classifier attention maps for different attribute editing operations, where we see that the classifier is capable of attending to the relevant regions with the learned attention-pooling layer. This component plays a vital role in removing the manual region masking that is necessary in \textit{blended diffusion} \cite{avrahami2022blended}, rendering our manipulation pipeline simple and compact. 

\subsection{Ablations}

Classifier-guidance is the key component in our framework that injects conditioning information (target attribute) to the diffusion process. In \cref{fig:acc_abl}, we illustrate how the capacity of the classifier affects the editing process. We set blended diffusion as the baseline, as CLIP guidance can be regarded as the coarsest level of guidance. Then we present three classifier variants with different classification accurac\nj{ies}. We observe that the classification performance clearly affects the spatial attention quality and the final editing performance, highlighting the importance of effective classifier finetuning. 

\begin{figure}[t]
\centering
  \includegraphics[width=0.96\columnwidth]{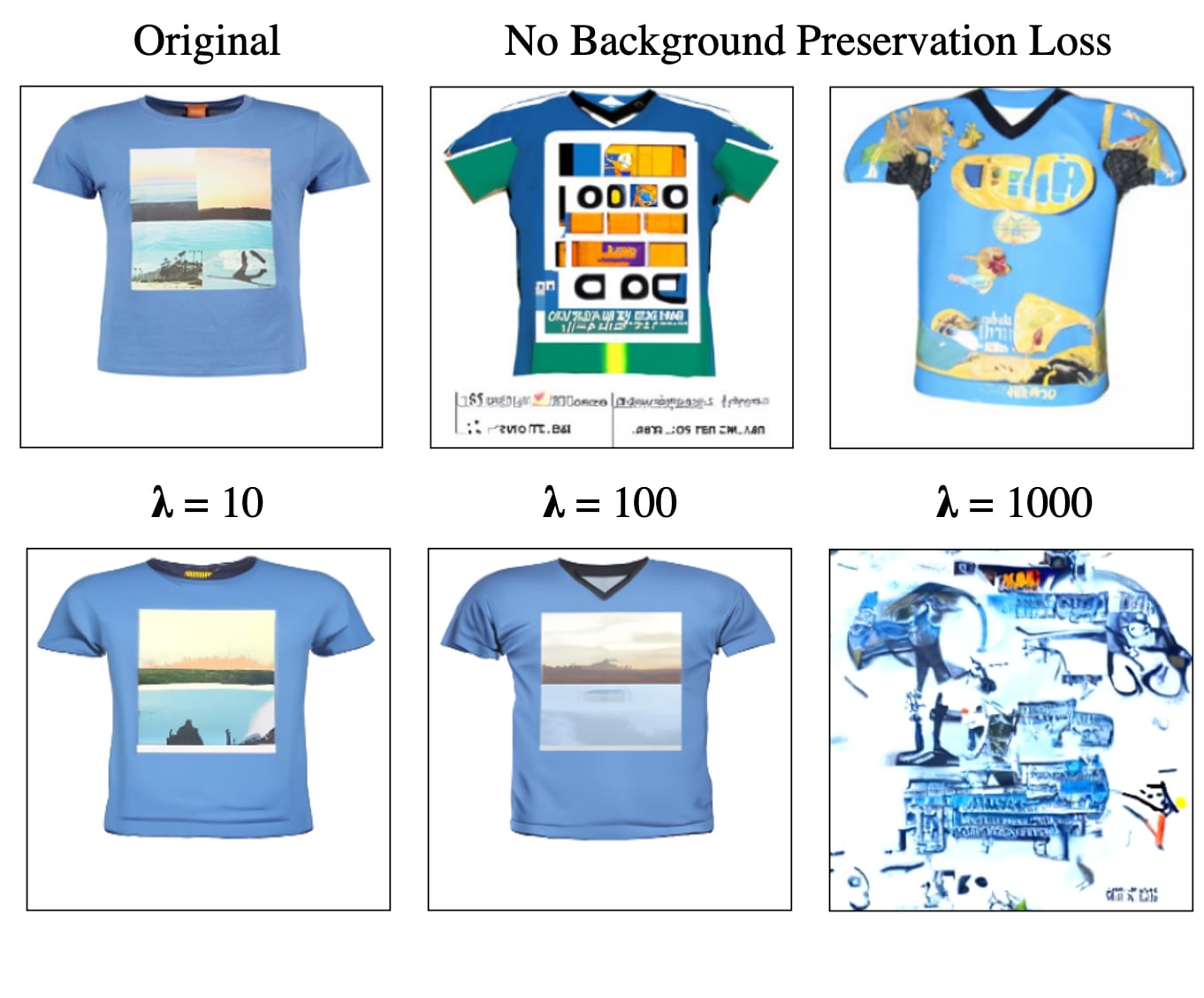}
  \caption{Ablation on loss components. We set ``v-neck collar'' as the target attribute. $\lambda$ refers to the weight multiplied to our classifier attribute guidance during the diffusion process.}
  \label{fig:hp_abl}
\end{figure}

\cref{fig:hp_abl} shows qualitative ablations on the core loss components and their hyperparameters. From the first row, we observe that background preservation loss is crucial for our framework, as there is no separate spatial mask that explicitly marks the region to be edited. Hence, without this term, the diffusion model generates random images that satisfy the given condition, \textit{e.g.,} random pictures of v-neck t-shirts.

The second row illustrates how the guidance weight affects the sample quality. With small guidance scale (low $\lambda$ value), we obtain realistic samples that are \textit{not} aligned with the attribute condition. When the guidance scale is too big, the diffusion process yields unsatisfactory outputs, possibly due to the train-test mismatch pointed out by \cite{saharia2022photorealistic}. We find the sweet spot at around $\lambda = 100$, where a good balance between the fidelity and the alignment is achieved. Hence, all of our experiments are conducted in this setting. 

\section{Conclusion}

In this paper, we have explored the prominent task of fashion attribute editing. As conventional approaches fall short in terms of scalability, we propose classifier-guided diffusion as a simple yet effective alternative. 
To that end, we train a multi-attribute classifier equipped with attention-pooling layer, and use its signal to guide the diffusion process. Empirical validations demonstrate the effectiveness of our framework as a generic attribute editing pipeline.

{\small
\bibliographystyle{ieee_fullname}
\bibliography{egbib}

\begin{thebibliography}{10}\itemsep=-1pt

\bibitem{ak2018efficient}
Kenan~E Ak, Joo~Hwee Lim, Jo~Yew Tham, and Ashraf~A Kassim.
\newblock Efficient multi-attribute similarity learning towards attribute-based
  fashion search.
\newblock In {\em 2018 IEEE Winter Conference on Applications of Computer
  Vision (WACV)}, pages 1671--1679. IEEE, 2018.

\bibitem{ak2019attribute}
Kenan~E Ak, Joo~Hwee Lim, Jo~Yew Tham, and Ashraf~A Kassim.
\newblock Attribute manipulation generative adversarial networks for fashion
  images.
\newblock In {\em Proceedings of the IEEE/CVF International Conference on
  Computer Vision}, pages 10541--10550, 2019.

\bibitem{avrahami2022blended}
Omri Avrahami, Dani Lischinski, and Ohad Fried.
\newblock Blended diffusion for text-driven editing of natural images.
\newblock In {\em Proceedings of the IEEE/CVF Conference on Computer Vision and
  Pattern Recognition}, pages 18208--18218, 2022.

\bibitem{chen2021empirical}
Xinlei Chen, Saining Xie, and Kaiming He.
\newblock An empirical study of training self-supervised vision transformers.
\newblock In {\em Proceedings of the IEEE/CVF International Conference on
  Computer Vision}, pages 9640--9649, 2021.

\bibitem{choi2018stargan}
Yunjey Choi, Minje Choi, Munyoung Kim, Jung-Woo Ha, Sunghun Kim, and Jaegul
  Choo.
\newblock Stargan: Unified generative adversarial networks for multi-domain
  image-to-image translation.
\newblock In {\em Proceedings of the IEEE conference on computer vision and
  pattern recognition}, pages 8789--8797, 2018.

\bibitem{deng2009imagenet}
Jia Deng, Wei Dong, Richard Socher, Li-Jia Li, Kai Li, and Li Fei-Fei.
\newblock Imagenet: A large-scale hierarchical image database.
\newblock In {\em 2009 IEEE conference on computer vision and pattern
  recognition}, pages 248--255. Ieee, 2009.

\bibitem{dhariwal2021diffusion}
Prafulla Dhariwal and Alexander Nichol.
\newblock Diffusion models beat gans on image synthesis.
\newblock {\em Advances in Neural Information Processing Systems},
  34:8780--8794, 2021.

\bibitem{dosovitskiy2020image}
Alexey Dosovitskiy, Lucas Beyer, Alexander Kolesnikov, Dirk Weissenborn,
  Xiaohua Zhai, Thomas Unterthiner, Mostafa Dehghani, Matthias Minderer, Georg
  Heigold, Sylvain Gelly, et~al.
\newblock An image is worth 16x16 words: Transformers for image recognition at
  scale.
\newblock {\em arXiv preprint arXiv:2010.11929}, 2020.

\bibitem{he2019attgan}
Zhenliang He, Wangmeng Zuo, Meina Kan, Shiguang Shan, and Xilin Chen.
\newblock Attgan: Facial attribute editing by only changing what you want.
\newblock {\em IEEE transactions on image processing}, 28(11):5464--5478, 2019.

\bibitem{hertz2022prompt}
Amir Hertz, Ron Mokady, Jay Tenenbaum, Kfir Aberman, Yael Pritch, and Daniel
  Cohen-Or.
\newblock Prompt-to-prompt image editing with cross attention control.
\newblock {\em arXiv preprint arXiv:2208.01626}, 2022.

\bibitem{ho2020denoising}
Jonathan Ho, Ajay Jain, and Pieter Abbeel.
\newblock Denoising diffusion probabilistic models.
\newblock {\em Advances in Neural Information Processing Systems},
  33:6840--6851, 2020.

\bibitem{ho2022classifier}
Jonathan Ho and Tim Salimans.
\newblock Classifier-free diffusion guidance.
\newblock {\em arXiv preprint arXiv:2207.12598}, 2022.

\bibitem{ho2022video}
Jonathan Ho, Tim Salimans, Alexey Gritsenko, William Chan, Mohammad Norouzi,
  and David~J Fleet.
\newblock Video diffusion models.
\newblock {\em arXiv preprint arXiv:2204.03458}, 2022.

\bibitem{karras2019style}
Tero Karras, Samuli Laine, and Timo Aila.
\newblock A style-based generator architecture for generative adversarial
  networks.
\newblock In {\em Proceedings of the IEEE/CVF conference on computer vision and
  pattern recognition}, pages 4401--4410, 2019.

\bibitem{kim2021diffusionclip}
Gwanghyun Kim and Jong~Chul Ye.
\newblock Diffusionclip: Text-guided image manipulation using diffusion models.
\newblock 2021.

\bibitem{kim2019u}
Junho Kim, Minjae Kim, Hyeonwoo Kang, and Kwanghee Lee.
\newblock U-gat-it: Unsupervised generative attentional networks with adaptive
  layer-instance normalization for image-to-image translation.
\newblock {\em arXiv preprint arXiv:1907.10830}, 2019.

\bibitem{kong2021smoothing}
Chaerin Kong, Jeesoo Kim, Donghoon Han, and Nojun Kwak.
\newblock Smoothing the generative latent space with mixup-based distance
  learning.
\newblock {\em arXiv preprint arXiv:2111.11672}, 2021.

\bibitem{kwon2022tailor}
Youngjoong Kwon, Stefano Petrangeli, Dahun Kim, Haoliang Wang, Viswanathan
  Swaminathan, and Henry Fuchs.
\newblock Tailor me: An editing network for fashion attribute shape
  manipulation.
\newblock In {\em Proceedings of the IEEE/CVF Winter Conference on Applications
  of Computer Vision}, pages 3831--3840, 2022.

\bibitem{liu2021swin}
Ze Liu, Yutong Lin, Yue Cao, Han Hu, Yixuan Wei, Zheng Zhang, Stephen Lin, and
  Baining Guo.
\newblock Swin transformer: Hierarchical vision transformer using shifted
  windows.
\newblock In {\em Proceedings of the IEEE/CVF International Conference on
  Computer Vision}, pages 10012--10022, 2021.

\bibitem{lugmayr2022repaint}
Andreas Lugmayr, Martin Danelljan, Andres Romero, Fisher Yu, Radu Timofte, and
  Luc Van~Gool.
\newblock Repaint: Inpainting using denoising diffusion probabilistic models.
\newblock In {\em Proceedings of the IEEE/CVF Conference on Computer Vision and
  Pattern Recognition}, pages 11461--11471, 2022.

\bibitem{nichol2021glide}
Alex Nichol, Prafulla Dhariwal, Aditya Ramesh, Pranav Shyam, Pamela Mishkin,
  Bob McGrew, Ilya Sutskever, and Mark Chen.
\newblock Glide: Towards photorealistic image generation and editing with
  text-guided diffusion models.
\newblock {\em arXiv preprint arXiv:2112.10741}, 2021.

\bibitem{nichol2021improved}
Alexander~Quinn Nichol and Prafulla Dhariwal.
\newblock Improved denoising diffusion probabilistic models.
\newblock In {\em International Conference on Machine Learning}, pages
  8162--8171. PMLR, 2021.

\bibitem{patashnik2021styleclip}
Or Patashnik, Zongze Wu, Eli Shechtman, Daniel Cohen-Or, and Dani Lischinski.
\newblock Styleclip: Text-driven manipulation of stylegan imagery.
\newblock In {\em Proceedings of the IEEE/CVF International Conference on
  Computer Vision}, pages 2085--2094, 2021.

\bibitem{ping2019fashion}
Qing Ping, Bing Wu, Wanying Ding, and Jiangbo Yuan.
\newblock Fashion-attgan: Attribute-aware fashion editing with multi-objective
  gan.
\newblock In {\em Proceedings of the IEEE/CVF conference on computer vision and
  pattern recognition workshops}, pages 0--0, 2019.

\bibitem{radford2021learning}
Alec Radford, Jong~Wook Kim, Chris Hallacy, Aditya Ramesh, Gabriel Goh,
  Sandhini Agarwal, Girish Sastry, Amanda Askell, Pamela Mishkin, Jack Clark,
  et~al.
\newblock Learning transferable visual models from natural language
  supervision.
\newblock In {\em International Conference on Machine Learning}, pages
  8748--8763. PMLR, 2021.

\bibitem{ramesh2022hierarchical}
Aditya Ramesh, Prafulla Dhariwal, Alex Nichol, Casey Chu, and Mark Chen.
\newblock Hierarchical text-conditional image generation with clip latents.
\newblock {\em arXiv preprint arXiv:2204.06125}, 2022.

\bibitem{richardson2021encoding}
Elad Richardson, Yuval Alaluf, Or Patashnik, Yotam Nitzan, Yaniv Azar, Stav
  Shapiro, and Daniel Cohen-Or.
\newblock Encoding in style: a stylegan encoder for image-to-image translation.
\newblock In {\em Proceedings of the IEEE/CVF conference on computer vision and
  pattern recognition}, pages 2287--2296, 2021.

\bibitem{rombach2022high}
Robin Rombach, Andreas Blattmann, Dominik Lorenz, Patrick Esser, and Bj{\"o}rn
  Ommer.
\newblock High-resolution image synthesis with latent diffusion models.
\newblock In {\em Proceedings of the IEEE/CVF Conference on Computer Vision and
  Pattern Recognition}, pages 10684--10695, 2022.

\bibitem{ronneberger2015u}
Olaf Ronneberger, Philipp Fischer, and Thomas Brox.
\newblock U-net: Convolutional networks for biomedical image segmentation.
\newblock In {\em International Conference on Medical image computing and
  computer-assisted intervention}, pages 234--241. Springer, 2015.

\bibitem{saharia2022photorealistic}
Chitwan Saharia, William Chan, Saurabh Saxena, Lala Li, Jay Whang, Emily
  Denton, Seyed Kamyar~Seyed Ghasemipour, Burcu~Karagol Ayan, S~Sara Mahdavi,
  Rapha~Gontijo Lopes, et~al.
\newblock Photorealistic text-to-image diffusion models with deep language
  understanding.
\newblock {\em arXiv preprint arXiv:2205.11487}, 2022.

\bibitem{selvaraju2017grad}
Ramprasaath~R Selvaraju, Michael Cogswell, Abhishek Das, Ramakrishna Vedantam,
  Devi Parikh, and Dhruv Batra.
\newblock Grad-cam: Visual explanations from deep networks via gradient-based
  localization.
\newblock In {\em Proceedings of the IEEE international conference on computer
  vision}, pages 618--626, 2017.

\bibitem{sohl2015deep}
Jascha Sohl-Dickstein, Eric Weiss, Niru Maheswaranathan, and Surya Ganguli.
\newblock Deep unsupervised learning using nonequilibrium thermodynamics.
\newblock In {\em International Conference on Machine Learning}, pages
  2256--2265. PMLR, 2015.

\bibitem{song2020denoising}
Jiaming Song, Chenlin Meng, and Stefano Ermon.
\newblock Denoising diffusion implicit models.
\newblock {\em arXiv preprint arXiv:2010.02502}, 2020.

\bibitem{steiner2021train}
Andreas Steiner, Alexander Kolesnikov, Xiaohua Zhai, Ross Wightman, Jakob
  Uszkoreit, and Lucas Beyer.
\newblock How to train your vit? data, augmentation, and regularization in
  vision transformers.
\newblock {\em arXiv preprint arXiv:2106.10270}, 2021.

\bibitem{tov2021designing}
Omer Tov, Yuval Alaluf, Yotam Nitzan, Or Patashnik, and Daniel Cohen-Or.
\newblock Designing an encoder for stylegan image manipulation.
\newblock {\em ACM Transactions on Graphics (TOG)}, 40(4):1--14, 2021.

\bibitem{yu2022coca}
Jiahui Yu, Zirui Wang, Vijay Vasudevan, Legg Yeung, Mojtaba Seyedhosseini, and
  Yonghui Wu.
\newblock Coca: Contrastive captioners are image-text foundation models.
\newblock {\em arXiv preprint arXiv:2205.01917}, 2022.

\bibitem{lpips}
R. Zhang, P. Isola, A.~A. Efros, E. Shechtman, and O. Wang.
\newblock The unreasonable effectiveness of deep features as a perceptual
  metric.
\newblock In {\em 2018 IEEE/CVF Conference on Computer Vision and Pattern
  Recognition (CVPR)}, pages 586--595, Los Alamitos, CA, USA, jun 2018. IEEE
  Computer Society.

\bibitem{zhou2016learning}
Bolei Zhou, Aditya Khosla, Agata Lapedriza, Aude Oliva, and Antonio Torralba.
\newblock Learning deep features for discriminative localization.
\newblock In {\em Proceedings of the IEEE conference on computer vision and
  pattern recognition}, pages 2921--2929, 2016.

\bibitem{zhu2017unpaired}
Jun-Yan Zhu, Taesung Park, Phillip Isola, and Alexei~A Efros.
\newblock Unpaired image-to-image translation using cycle-consistent
  adversarial networks.
\newblock In {\em Proceedings of the IEEE international conference on computer
  vision}, pages 2223--2232, 2017.

\end{thebibliography}
}

\end{document}